\documentclass[10pt,journal,compsoc,fleqn]{IEEEtran}

\ifCLASSOPTIONcompsoc
  \usepackage[nocompress]{cite}
\else
  \usepackage{cite}
\fi

\ifCLASSINFOpdf
\else
\fi

	\usepackage{graphicx}
	\usepackage{times,amsmath,epsfig}
	\usepackage{enumerate}
	\usepackage{graphics}
	\usepackage{balance}
	\usepackage{booktabs}
	\usepackage{multirow}
	\usepackage{amssymb}
	\usepackage{pifont}
	\usepackage{rotating}
	\usepackage{subfigure}
	\usepackage{amsmath} 
	\usepackage{tikz}
	\usepackage{pgfplots}
	\usepgfplotslibrary{groupplots}
	\usetikzlibrary{plotmarks}
	\usepgfplotslibrary{units}
	\usepackage{verbatim}
	
	\usepackage{relsize} 

	\usepackage{amsfonts}
	\usepackage{algorithm}
	\usepackage{algpseudocode}
	\algrenewcommand\alglinenumber[1]{\small #1:} 

	\algnewcommand{\IIf}[1]{\State\algorithmicif\ #1\ \algorithmicthen}
	\algnewcommand{\EndIIf}{\unskip\ \algorithmicend\ \algorithmicif}
	\algnewcommand{\FForAll}[1]{\State\algorithmicforall\ #1\ \algorithmicthen}
	\algnewcommand{\EndFFor}{\unskip\ \algorithmicend\ \algorithmicforall}

	\usepackage{footnote}
	
	\usepackage{dsfont}
	
	\usepackage{epstopdf}

	\definecolor{color1}{rgb}{0.545098, 0, 0.545098}
	\definecolor{color2}{rgb}{0.0980392, 0.0980392, 0.439216}
	\definecolor{color3}{rgb}{0.196078, 0.803922, 0.196078}
	\definecolor{color4}{rgb}{0.545098, 0, 0}

	\renewenvironment{thebibliography}[1]{%
		\begin{oldthebibliography}{#1}%
			\linespread{1}
			\small
			\setlength{\parskip}{0ex}%
			\setlength{\itemsep}{0ex}%
		}%
		{%
		\end{oldthebibliography}%
	}
	
	\usepackage{hyperref}
	\hypersetup{
		colorlinks=true,
		linkcolor=blue,
		filecolor=magenta,
		urlcolor=cyan,
	}
	
	\usepackage{cite}

	\usepackage[nomessages]{fp}
	
	\usepackage{scrextend}

	\newtheorem{definition}{Definition}
	\usepackage[utf8]{inputenc}
	\newcommand\xqed[1]{%
		\leavevmode\unskip\penalty9999 \hbox{}\nobreak\hfill
		\quad\hbox{#1}}
	\newcommand\DefBox{\xqed{$\Box$}}

	\usepackage{chngcntr}
	\counterwithin{corollary}{definition}
	
	\usepackage{flexisym}

	\usepackage{nicefrac}

	\algtext*{EndWhile} 
	\algtext*{EndFor} 
	\algtext*{EndIf}    

	\usepackage{ragged2e}

	\algdef{SE}[EVENT]{Event}{EndEvent}[1]{\textbf{upon receiving}\ #1\ \algorithmicdo}{\algorithmicend\ \textbf{event}}%
	\algtext*{EndEvent}

	\usepackage{epstopdf}

    \pgfplotsset{compat=1.16}

    \usepackage{mathtools}
	\DeclarePairedDelimiter\ceil{\lceil}{\rceil}

\begin{document}

    \makeatletter
    \def\ps@IEEEtitlepagestyle{%
        \def\@oddfoot{\mycopyrightnotice}%
        \def\@evenfoot{}%
    }
    \def\mycopyrightnotice{%
        {\hfill \footnotesize
        \begin{minipage}{\textwidth}
            \centering
            Copyright~\copyright~2021 IEEE. Personal use of this material is permitted. Permission from IEEE must be obtained for all other uses, in any current or future media, including reprinting/republishing this material for advertising or promotional purposes, creating new collective works, for resale or redistribution to servers or lists, or reuse of any copyrighted component of this work in other works.
        \end{minipage}
        }
    }
    \makeatother

    \title{AI Back-End as a Service for Learning Switching of Mobile Apps between the Fog and the Cloud}

	\author{Dionysis~Athanasopoulos~and~Dewei~Liu
		\IEEEcompsocitemizethanks{\IEEEcompsocthanksitem Dionysis Athanasopoulos and Dewei Liu are with the School of Electronics, Electrical Engineering \& Computer Science, Queen's University of Belfast, Northern Ireland, UK.
			E-mails: \{D.Athanasopoulos,dliu08\}@qub.ac.uk
		}
		\thanks{Manuscript received March 31, 2021; accepted October 1, 2021.}
	}

	\markboth{IEEE TRANSACTIONS ON SERVICES COMPUTING,~Vol.~XX, No.~XX, OCTOBER~2021}%
	{Shell \MakeLowercase{\textit{et al.}}: Bare Demo of IEEEtran.cls for Computer Society Journals}

\IEEEtitleabstractindextext{%
	\begin{abstract}
	\justifying
	Given that cloud servers are usually remotely located from the devices of mobile apps, the end-users of the apps can face delays. The Fog has been introduced to augment the apps with machines located at the network edge close to the end-users. However, edge machines are usually resource constrained. Thus, the execution of online data-analytics on edge machines may not be feasible if the time complexity of the data-analytics algorithm is high. To overcome this, multiple instances of the back-end should be deployed on edge and remote machines. In this case, the research question is how the switching of the app among the instances of the back-end can be dynamically decided based on the response time of the service instances. To answer this, we contribute an AI approach that trains machine-learning models of the response time of service instances. Our approach extends a back-end as a service into an AI self-back-end as a service that self-decides at runtime the right edge/remote instance that achieves the lowest response-time. We evaluate the accuracy and the efficiency of our approach by using real-word machine-learning datasets on an existing auction app.
		\end{abstract}
	
	}

	\maketitle

	\IEEEdisplaynontitleabstractindextext

	\IEEEpeerreviewmaketitle

\section{Introduction}
Mobile apps that perform online data-analytics are rapidly increasing.
Ofcom's Online UK Nation 2020 report states the percentage of users of mobile apps was significantly increased the last year\footnote{\url{https://www.ofcom.org.uk/__data/assets/pdf_file/0027/196407/online-nation-2020-report.pdf}}.
While we have high expectations from apps, mobile devices cannot meet these expectations due to their limited hardware-resources.
In this case, the back-end of apps usually run computationally demanding online data-analytics on powerful remote servers by adopting the Web-service technology (a.k.a., mobile back-end as a service\footnote{\url{https://azure.microsoft.com/en-us/solutions/mobile}}).
Data analytics refer to algorithms that mine knowledge from data (e.g., price predictions) \cite{DBLP:conf/caise/PlebaniGABCKPPT17}.

However, the use of remote servers for performing data analytics may lead mobile apps to experience delays in receiving the analysis output.
For instance, the network latency is not always predictable \cite{Aikat:2003:VTR:948205.948241} and the end-users of mobile apps can face high end-to-end response times.
The response time equals to the sum of the execution time of the analysis on a machine plus the latency of the underlying communication links.

The Fog has been introduced to reduce response times of mobile apps \cite{DBLP:journals/csur/HongV19}.
The Fog augments mobile apps with devices located near/at the network edge.
In this way, the Fog has arrived to mitigate the latency issues that emerge from the usage of remote machines.
Given edge devices are resource constrained compared to powerful remote servers, the execution of data analytics on edge devices is not always feasible for large datasets.
To overcome this, the execution of data analytics should combine edge and remote machines\footnote{We do not consider in this work the financial debt of hiring machines.} \cite{DBLP:conf/caise/PlebaniGABCKPPT17}.
In particular, multiple instances of the back-end should be deployed on both edge machines and remote machines.

The data analytics whose time complexity is low enough can be executed on edge machines.
On the contrary, data analytics whose time complexity is high cannot be executed on edge machines.
Thus, the \textit{dynamic switching} of the front-end (e.g., GUI) among the instances of the back-end should be decided.
However, the dynamic decision of the right instance is not straightforward because it depends on the dataset provided as input to the data-analysis algorithm.
Thus, the following research question is raised:

\textit{``How the switching among the instances of a back-end can be decided based on the response time of the back-end?''}

To answer this, the response time of a service back-end should get first calculated.
However, data analytics implement complex algorithms that contain many interconnected data-structures whose response time depends on the particulars of the algorithms' utility (a.k.a., the size of input datasets).
Moreover, the response time of a back-end further depends on the machines where the service instances have been deployed.
Overall, we focus in this work on the following empirical factors that affect the response time of service instances: input datasets and underlying machines\footnote{Please note we focus on the response time of service instances in this work (i.e., the round-trip time from the service client to the service instance) because we assume we do not have access to the service implementation and to the resources (e.g., CPU, RAM) consumed on the used machines. In this way, we cannot calculate performance metrics (like CPU, memory consumption) that are related to the performance of the service instances on the machines.}.

The problem is that the response time of service instances cannot be necessarily estimated at the development time of mobile apps because the input datasets of the back-end and the used machines may have not been decided at the development time of the apps.
Thus, the response time of service instances should be \textit{predicted at runtime} based on the input datasets and the available edge/remote machines.
If the predicted response-time is low, then the app should get bound to an edge instance of the back-end.
On the contrary, if the predicted response-time is high, then the app should get bound to a remote instance of the back-end.

We meet approaches in the literature that estimate the response time of services, without analysing the source code of the service implementation.
These approaches estimate the response time via analysing the Web API of a service (e.g., \cite{DBLP:conf/icsoc/AthanasopoulosM19}).
The restriction of the existing approaches is that they build mathematical expressions of the response time in a manner \textit{tied to the independent variables} of the expressions.
The independent variables usually include input/output data-types of Web APIs and/or the type of edge/remote machines.
Given that the construction manner of a mathematical expression is tied to its independent variables, a new mechanism should be developed for different independent variables.
Thus, we face the \textit{challenge} in proposing a generic approach that builds expressions without being tied to independent variables.

To address this challenge, we contribute an artificial intelligence (AI) approach that takes any independent variable as input and constructs machine-learning models that capture the response time of service instances.
We specify the architectural design and the algorithmic mechanisms of the AI approach.
The approach is applied to mobile apps that follow the architectural pattern of the mobile back-end as a service.
These apps usually include at least three interconnected components: the front-end, the controller, and the back-end.
Our approach extends a back-end as a service into an AI back-end as a service.
The AI back-end encapsulates the available edge/remote instances of the service back-end.
The AI back-end further constructs machine-learning regression expressions of the response time of the edge instances and the remote instances of the service back-end.

The machine-learning technique implemented by the AI back-end has been designed in a generic way to use multiple machine-learning models (e.g., k-nearest neighbours, support vector machine).
The AI back-end uses the trained machine-learning expressions to self-decide at runtime the edge/remote instance that achieves the lowest response-time.
In other words, our approach further extends the AI back-end to an AI self-back-end.

We conduct a set of experiments where multiple machine-learning models were implemented.
We compare and contrast the usage of four machine-learning models (neural networks, k-nearest neighbours, support vector machines, decision trees).
We evaluate the accuracy of our approach via measuring the percentage of the correct switching-decisions made by the AI self-back-end.
We further evaluate the efficiency of the approach via measuring the improvement in the response time of the app on all the switching decisions.
We use in our experiments real-word datasets collected from UC Irvine machine-learning repository\footnote{\url{http://archive.ics.uci.edu/ml/index.php}}.
We also use a real-world auction mobile app as a case-study.

The rest of the paper is structured as follows.
Section \ref{sec:SOTA} categorises and compares the related approaches.
Section \ref{sec:app} defines the conceptual model of mobile apps with back-end as a service.
Section \ref{sec:case-study} describes the case study that we use as a running example in the paper.
Section \ref{sec:infrastructure} defines the Fog infrastructure that we adopt in this work.
Section \ref{sec:autonomousService} specifies the architecture of the AI self-back-end as a service.
Section \ref{sec:loop} details the mechanisms of the AI self-back-end.
Section \ref{sec:evaluation} presents the experimental evaluation of our approach.
Finally, Section \ref{sec:ConclusionsAndFutureWork} summarises our contribution and discusses the future research-directions.

\section{Related Work}\label{sec:SOTA}
	\begin{table*}
		\begin{center}
			\caption{Categorization and comparison of the existing approaches that deploy/execute mobile apps via the Fog. \label{Table:sota}} 
			\scalebox{1.0}{
				\begin{tabular}{ c c c| c c }
					\cline{2-5}
					\multirow{18}{*}{\rotatebox[origin=c]{90}{\textbf{Runtime of apps}}} & \multicolumn{1}{|l}{\textit{Non-functional attribute:}} & \multicolumn{1}{l|}{latency \cite{Saurez:2016:IDM:2933267.2933317}, CPU \cite{DBLP:journals/fgcs/EvangelidisPB18,DBLP:conf/wosp/BarnaLFSW18,DBLP:conf/icse/BarnaKFL17}} & \multicolumn{1}{|l}{\textit{Performance indicator:}} & \multicolumn{1}{l|}{power consumption \cite{DBLP:conf/icfec/BrogiFI17,7511147}} \\
					& \multicolumn{1}{|l}{} & \multicolumn{1}{l|}{response time \cite{DBLP:journals/tse/ChenB17,DBLP:conf/wosp/BarnaLFSW18,DBLP:conf/icse/BarnaKFL17,DBLP:conf/icsoc/AthanasopoulosM19}} & \multicolumn{1}{l}{} & \multicolumn{1}{l|}{service delay \cite{8029252,DBLP:journals/iotj/YousefpourPIKWC19}, resource consumption \cite{DBLP:journals/iotj/YousefpourPIKWC19}} \\
					& \multicolumn{1}{|l}{} & \multicolumn{1}{l|}{throughput \cite{DBLP:journals/tse/ChenB17}} & \multicolumn{1}{l}{} & \multicolumn{1}{l|}{computation cost \cite{DBLP:journals/computing/TranTPKHH17}, energy consumption \cite{DBLP:journals/tsc/ZhangLYCKL19,DBLP:journals/tsc/ChenWL19}} \\
					& \multicolumn{1}{|l}{} & \multicolumn{1}{l|}{availability \cite{DBLP:journals/tse/ChenB17}, reliability \cite{DBLP:journals/tse/ChenB17}} & \multicolumn{1}{l}{} & \multicolumn{1}{l|}{fog-to-cloud bandwidth \cite{DBLP:journals/iotj/YousefpourPIKWC19}, latency \cite{DBLP:journals/tsc/WangVMN20}} \\
					
					& \multicolumn{1}{|l}{\multirow{1}{*}{\textit{Model/Execution}}} & \multicolumn{1}{l|}{tree traversal \cite{Saurez:2016:IDM:2933267.2933317}} & \multicolumn{1}{|l}{\textit{Model/Execution}} & \multicolumn{1}{l|}{sub-graph isomorphism \cite{DBLP:conf/icfec/BrogiFI17}} \\
					& \multicolumn{1}{|l}{\textit{plan generation:}} & \multicolumn{1}{l|}{machine learning \cite{DBLP:journals/tse/ChenB17,DBLP:conf/icsoc/AthanasopoulosM19}} & \multicolumn{1}{|l}{\textit{plan generation:}} & \multicolumn{1}{l|}{graph traversal \cite{8029252}, deep learning \cite{DBLP:journals/tsc/ZhangLYCKL19}} \\
					& \multicolumn{1}{|l}{} & \multicolumn{1}{l|}{Markov chains \cite{DBLP:journals/fgcs/EvangelidisPB18}} & \multicolumn{1}{|l}{} & \multicolumn{1}{l|}{indexing data-structure \cite{7511147}} \\
					& \multicolumn{1}{|l}{} & \multicolumn{1}{l|}{layered queuing models \cite{DBLP:conf/wosp/BarnaLFSW18,DBLP:conf/icse/BarnaKFL17}} & \multicolumn{1}{|l}{} & \multicolumn{1}{l|}{bipartite matching \cite{DBLP:journals/computing/TranTPKHH17}, maximum independent set \cite{DBLP:journals/tsc/ChenWL19}} \\
					& \multicolumn{1}{|l}{} & \multicolumn{1}{l|}{} & & \multicolumn{1}{l|}{integer non-linear programming \cite{DBLP:journals/iotj/YousefpourPIKWC19}} \\
					& \multicolumn{1}{|l}{\textit{Model/plan}} & \multicolumn{1}{l|}{control knobs \cite{DBLP:journals/tse/ChenB17}, input data \cite{DBLP:conf/icsoc/AthanasopoulosM19}} & \multicolumn{1}{|l}{\textit{Model/plan}} & \multicolumn{1}{l|}{discrete time steps \cite{8029252,7511147}} \\
					& \multicolumn{1}{|l}{\textit{in function of}:} & \multicolumn{1}{l|}{service requests \cite{DBLP:journals/tse/ChenB17,DBLP:conf/wosp/BarnaLFSW18,DBLP:conf/icse/BarnaKFL17}} & \multicolumn{1}{l}{\textit{in function of}:} & \multicolumn{1}{l|}{incoming traffic rate \cite{DBLP:journals/iotj/YousefpourPIKWC19}} \\
					& \multicolumn{1}{|l}{} & \multicolumn{1}{l|}{discrete time steps \cite{DBLP:journals/fgcs/EvangelidisPB18,DBLP:conf/wosp/BarnaLFSW18,DBLP:conf/icse/BarnaKFL17}} & & \multicolumn{1}{l|}{energy consumption \cite{DBLP:journals/tsc/ZhangLYCKL19}, service requests \cite{DBLP:journals/tsc/WangVMN20}} \\
					& \multicolumn{1}{|l}{} & \multicolumn{1}{l|}{input data size \cite{DBLP:conf/wosp/BarnaLFSW18,DBLP:conf/icse/BarnaKFL17}} & & \multicolumn{1}{l|}{multiple tasks of multiple users \cite{DBLP:journals/tsc/ChenWL19}} \\
					& \multicolumn{2}{|c|}{\multirow{2}{*}{$\textbf{Q}_{\textbf{3}}$}} & \multicolumn{2}{|c|}{\multirow{2}{*}{$\textbf{Q}_{\textbf{4}}$}} \\
					& \multicolumn{2}{|c|}{} & \multicolumn{2}{|c|}{} \\ \cline{2-5}
					
					\multirow{6}{*}{\rotatebox[origin=c]{90}{\textbf{Deployment}}} & \multicolumn{1}{|l}{\textit{Non-functional attribute}:} & \multicolumn{1}{l|}{latency \cite{DBLP:journals/iotj/BrogiF17}} & \multicolumn{1}{|l}{\textit{Performance indicator:}} & \multicolumn{1}{l|}{power consumption \cite{DBLP:conf/icfec/BrogiFI17,7467406,SPE:SPE2509}} \\
					& \multicolumn{1}{|l}{} & \multicolumn{1}{l|}{bandwidth \cite{DBLP:journals/iotj/BrogiF17}, response time \cite{7467406}} & \multicolumn{1}{l}{} & \multicolumn{1}{l|}{computation \& bandwidth resources \cite{DBLP:journals/tsc/MaGWJG20}} \\
					& \multicolumn{1}{|l}{\textit{Deployment plan}} & \multicolumn{1}{l|}{subgraph isomorphism \cite{DBLP:journals/iotj/BrogiF17}} & \multicolumn{1}{|l}{\textit{Deployment plan}} & \multicolumn{1}{l|}{subgraph isomorphism \cite{DBLP:conf/icfec/BrogiFI17}, integer programming \cite{7467406}} \\
					& \multicolumn{1}{|l}{\textit{generation:}} & \multicolumn{1}{c}{} & \multicolumn{1}{|l}{\textit{generation:}} & \multicolumn{1}{l|}{graph traversal \cite{SPE:SPE2509}, Game theory \cite{DBLP:journals/tsc/MaGWJG20}} \\
					& \multicolumn{2}{|c|}{\multirow{2}{*}{$\textbf{Q}_{\textbf{1}}$}} & \multicolumn{2}{|c|}{\multirow{2}{*}{$\textbf{Q}_{\textbf{2}}$}} \\
					& \multicolumn{2}{|c|}{} & \multicolumn{2}{|c|}{} \\ \cline{2-5}
					& \multicolumn{2}{c}{\multirow{2}{*}{\textbf{Non-functional requirements for apps}}} & \multicolumn{2}{c}{\multirow{2}{*}{\textbf{Runtime performance of Fog infrastructure}}}
				\end{tabular}
			}
		\end{center}
	\end{table*}

The existing approaches for executing mobile apps via the Fog generate deployment/execution plans to map instances of back-ends to edge/remote machines.
These plans aim at meeting the non-functional requirements for apps and/or at considering the runtime performance of apps and of the underlying Fog infrastructure.
For instance, back-ends are mapped to machines only if the requirements for the back-ends (e.g., CPU requirements) can be met by the machine characteristics (e.g., CPU cores).

We categorise the existing approaches in two dimensions: (i) non-functional requirements vs. runtime performance; (ii) deployment time vs. runtime of apps.
We form a categorisation matrix of four quadrants, $Q_1$-$Q_4$ (Table \ref{Table:sota}), that are detailed in Sections \ref{sec:quadrant1}-\ref{sec:quadrant4}.
Our approach falls into $Q_3$.
We finally provide in Section \ref{sec:comparison} an overall comparison of the existing approaches against our approach to highlight the research gap in the literature.

\subsection{Deployment Time of Mobile Apps}

\subsubsection{Non-Functional Requirements for Mobile Apps}\label{sec:quadrant1}
\cite{DBLP:journals/iotj/BrogiF17} proposes an automated mechanism for generating deployment plans that meet the requirements for the latency and the bandwidth of communication links.
\cite{DBLP:journals/iotj/BrogiF17} proposes a heuristic to get approximate solutions because the mapping of many back-ends to many machines is NP-hard (subgraph isomorphism problem).
\cite{7467406} proposes an automated mechanism for generating deployment plans that minimise the delay of a Fog infrastructure to serve requests to apps.
To approximate optimal deployment plans, \cite{7467406} adopts a mixed integer non-linear programming technique.

\subsubsection{Runtime Performance of Fog Infrastructure}\label{sec:quadrant2}
\cite{DBLP:conf/icfec/BrogiFI17,SPE:SPE2509} generates deployment plans that reduce the power consumption of apps on edge machines.
\cite{7467406} produces deployment plans that consider both the service delay of a Fog infrastructure and the power consumption of apps on edge/remote machines.
\cite{7467406} approximates optimal deployment plans by balancing the service time and the power consumption.
\cite{DBLP:journals/tsc/MaGWJG20} allocates service instances to edge machines to optimize the consumption of computation and bandwidth
resources on the machines.

\subsection{Runtime of Mobile Apps}

\subsubsection{Non-Functional Requirements for Mobile Apps}\label{sec:quadrant3}
\cite{Saurez:2016:IDM:2933267.2933317} proposes an approach for the migration a back-end from an edge machine to another if the response time of the back-end exceeds a time threshold.
\cite{DBLP:journals/tse/ChenB17} presents an approach for the machine-learning modelling of the runtime performance of apps in function of (software or hardware) control knobs and the number of the service requests.
\cite{DBLP:journals/fgcs/EvangelidisPB18} models and verifies (using discrete-time Markov chains) rule-based auto-scaling policies for cloud-based apps in function of the discrete elapsed time.
\cite{DBLP:conf/wosp/BarnaLFSW18,DBLP:conf/icse/BarnaKFL17} propose approaches for modelling (using Kalman filters) the runtime performance of apps in function of the input data, the number of the service requests, and the discrete elapsed time.
\cite{DBLP:conf/icsoc/AthanasopoulosM19} contributes the conceptual model and the algorithmic mechanisms of an autonomous approach \cite{DBLP:journals/computer/KephartC03}.
\cite{DBLP:conf/icsoc/AthanasopoulosM19} predicts the response time of a back-end by building predictive mathematical models.

\subsubsection{Runtime Performance of Fog Infrastructure}\label{sec:quadrant4}
\cite{8029252} proposes an analytical model for measuring the service delay of a Fog infrastructure in function of the elapsed discrete-time.
Based on this model, \cite{8029252} offloads computation from an edge machine to another in an online manner.
\cite{7511147} proposes an indexing of edge machines with respect to the trade-off that exist between the service delay and the power consumption.
\cite{DBLP:journals/computing/TranTPKHH17} proposes an offloading mechanism for migrating back-ends based on the end-users' budgets.
\cite{DBLP:journals/iotj/YousefpourPIKWC19} proposes a framework to decrease the service delay of a Fog infrastructure, the Fog-to-Cloud bandwidth, along with the resource usage on edge machines.
\cite{DBLP:journals/tsc/ZhangLYCKL19} proposes a scheduling model to reduce the energy consumed by dynamic switching activities between edge/cloud machines.
\cite{DBLP:journals/tsc/ChenWL19} proposes a multi-user multi-task offloading approach for the mobile edge cloud towards minimising the energy consumption on the edge/cloud machines.
\cite{DBLP:journals/tsc/WangVMN20} proposes mechanisms for provisioning and auto-scaling resources on edge machines.

\subsection{Comparing the Related Approaches}\label{sec:comparison}
Only two approaches monitor the app execution to regenerate deployment plans \cite{8029252,Saurez:2016:IDM:2933267.2933317}.
However, these approaches are reactive (i.e., they suspend the app execution) and lay between apps and operating systems (e.g., redeployment engines).
Contrarily, our approach runs at the application layer and pro-actively (without suspending the app execution) self-decides what instance of the back-end will be invoked.
Our past work proposed in \cite{DBLP:conf/icsoc/AthanasopoulosM19} is a pro-active approach too.
However, \cite{DBLP:conf/icsoc/AthanasopoulosM19} builds mathematical expressions of response time in a manner tied to the independent variables of the expressions.
On the contrary, our current work is a generic approach that builds machine-learning models in a manner that is not tied to independent variables.

\section{Mobile Back-End as a Service}\label{sec:app}
Mobile apps that follow the architectural pattern of the mobile back-end as a service usually include at least three interconnected components: the front-end, the controller, and the back-end.
The front-end of an app constitutes the part of the app that interacts with the end-users of the app (e.g., interaction via using a GUI).
The back-end of an app is provided as a service and is mainly responsible for data analysis\footnote{The storage of the back-end, along with the synchronisation of the data stored across multiple back-end instances, are beyond the scope of the paper.}.
Each service exposes a public Web API that is formally defined in Section \ref{sec:web-api}.
A controller facilitates the interaction between the front-end and the back-end.
The controller receives data from the front-end and forwards them for analysis to the back-end via using programming clients of the back-end to invoke service operations of the back-end.

\begin{figure}[t]
	\centering
	\includegraphics[scale=0.75]{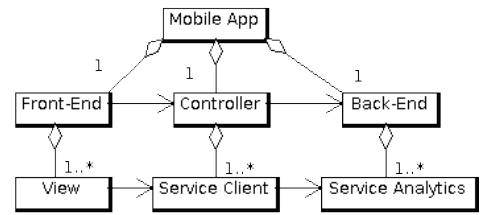}\\ 
	\caption{Mobile apps with (data-analytics) back-end as a service.}
	\label{fig:applicationModel}
\end{figure}

The architecture of mobile apps that we just described is depicted in Fig. \ref{fig:applicationModel}.
In particular, Fig. \ref{fig:applicationModel} visualises the architecture of mobile apps by adopting the notation of the UML class-diagram\footnote{\url{https://www.omg.org/spec/UML}}.
The UML class-diagram is a graphical notation used to construct and visualise the classes of a system and the relationships among objects of the classes\footnote{The arrows/links that we use in the UML diagrams of this paper represent the UML relationships of the association, the aggregation, and the composition.
Association exists when a class uses references to object(s) of another class and is visualised by an arrow.
Aggregation is a ``has a'' relationship between a composite class and its owned objects.
Aggregation is visualised by using a hollow diamond shape on the composite class.
Composition is a ``has a'' relationship indicating that when the composite object is finalised, its owned objects are also finalised.
Composition is visualised by using a filled diamond shape on the composite class.
The arrow/link of a UML relationship can be annotated with a number or an expression that corresponds to the multiplicity/cardinality of the relationship.
In particular, the expression of ``1'' denotes that a class uses exactly one object of another class.
The expression of ``1..*'' denotes that a class uses one or many objects of another class.}.

\subsection{Web API of Service Back-End}\label{sec:web-api}
The Web APIs of service back-ends can be specified via using the RESTful \cite{Richardson:2007:RWS:1406352} or the SOAP \cite{Thomas16} Web-service technologies.
To model the Web API of service back-ends independently of the underlying technology and language, we define the Web API in the following generic way.

\begin{definition}[Web API of service back-end]\label{def:api}
\textit{We define the Web API of a service back-end by the following tuple:}
\begin{equation}api = (n,\ ops,\ s,\ c)\end{equation}
\textit{The tuple comprises:}
	\begin{itemize}
		\item \textit{the name, $n$, of the service}
		\item \textit{the set, $ops=\{op_i\}$, of the operations of the service that correspond to the public operations of the service}
		\item \textit{the technology style, $s$ (RESTful or SOAP), of the Web API}
		\item \textit{the credentials, $c$ (e.g., JSON document), usually required by remote servers for authenticating the client identity.}
		\DefBox
	\end{itemize}
\end{definition}

\begin{definition}[Web API operation]\label{def:operation}
\textit{We define an operation $op$ of the Web API of a service by the following tuple:}
\begin{equation}op = (n, in, out)\end{equation}
\textit{The tuple consists of the name, $n$, of the operation, the set, $in=\{p_i\}$, of the input parameters, and the set, $out=\{p_i\}$, of the output parameters.} \DefBox
\end{definition}

An input/output parameter of an operation of a Web API can be a primitive parameter (Def. \ref{def:primitive-parameter}) or a complex parameter (Def. \ref{def:complex-parameter}).

\begin{definition}[Primitive parameter of an operation]\label{def:primitive-parameter}
\textit{We define a primitive parameter, $p$, of an operation by the following tuple:}
\begin{equation}p = (n, t)\end{equation}
\textit{The tuple comprises the name, $n$, of the parameter and the built-in data-type, $t$, of the parameter (e.g., \texttt{\small int}).} \DefBox
\end{definition}

\begin{definition}[Complex parameter of an operation]\label{def:complex-parameter}
\textit{We define a complex parameter, $p$, of an operation by the following tuple:}
\begin{equation}p = (n, g, t)\end{equation}
\textit{The tuple comprises the name, $n$, of the parameter, the grouping data-structure, $g$, of $t$ (e.g., a list), and the data-type of the (possibly absent) nested (primitive/complex) parameter, $t$.} \DefBox
\end{definition}

\section{Case Study}\label{sec:case-study}
Emma is an avid eBay user, always ready to snap up an auction bargain when she sees one.
But Emma faces a difficulty on deciding the best end-price for the auction to maximise her profits.
To this end, Emma envisions to use an auction app that can predict end-prices based on past bidding data in a timely manner \cite{10.1007/978-3-642-27609-5_16}.
Thus, Emma asked by Tom, who is a developer of mobile apps, to build an auction app for her.

\textbf{Mobile back-end as a service.}
Tom built an app in which the back-end is deployed as a service on a cloud machine (2.2GHz vCPU, 3.75GB RAM).
The data analysis performed by the back-end uses the k-means clustering algorithm to forecast the end-price of an online auction \cite{DataMiningBook06}.
In particular, the algorithm of the data analysis organises the past auction prices into groups of similar prices and further employs regression trees to predict end-prices.

\begin{figure}
    \centering
	\includegraphics[scale=0.65]{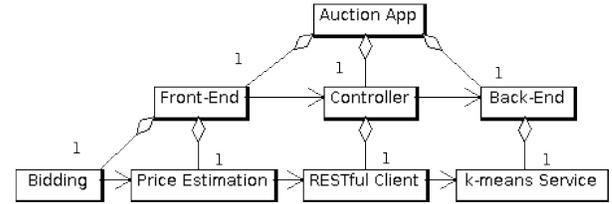}\\ 
	\caption{The architecture of the auction app with back-end as a service.}
	\label{fig:auctionModel}
\end{figure}

\textbf{Architecture of the auction app.}
The architecture of the auction app is depicted in Fig. \ref{fig:auctionModel}.
According to Fig. \ref{fig:auctionModel}, the front-end of the app includes two classes: the \texttt{\small Bidding} class and the \texttt{\small Price Estimation} class.
The \texttt{\small Bidding} class uses the \texttt{\small Price Estimation} class to predict the end-price of the current auction.
Following, the \texttt{\small Price Estimation} class interacts with the controller of the app and especially, the \texttt{\small Price Estimation} class interacts with the client of the RESTFul Web API of the service back-end of the app.
The service client invokes the Web API of the \texttt{\small k-means service} of the back-end via establishing an HTTP connection to the endpoint of the instance of the \texttt{\small k-means service} that has been deployed on a cloud machine.

\begin{figure}
	\centering
	\includegraphics[scale=0.9]{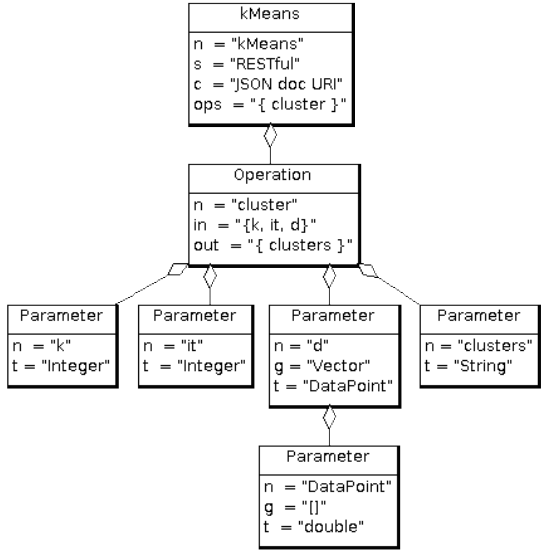}\\ 
	\caption{The Web API of the k-means service of the auction back-end.}
	\label{fig:kMeansAPI}
\end{figure}

\textbf{Web API of the auction back-end.}
Regarding the RESTful Web API of the \texttt{\small k-means service}, the Web API provides the single operation, \texttt{\small cluster}.
The parameters of this operation are the following: the clusters number, $k$, the clustering iterations, $it$, and a bidding dataset, $d$.
The dataset is specified as a set of multi-dimensional data-points.
In other words, the parameter of a bidding dataset is a complex parameter (Def. \ref{def:complex-parameter}) that consists of the Java grouping data-structure, \texttt{\small Vector}, and further includes the nested complex parameter, \texttt{\small DataPoint}.
The \texttt{\small DataPoint} parameter consists of the array data-structure that contains \texttt{\small double} numbers (a.k.a., \texttt{\small double[]}).
Overall, the Web API of the \texttt{\small k-means service} (Web API, operation, and parameters) is depicted in Fig. \ref{fig:kMeansAPI} by using the definition of the Web API of a service back-end that was provided in Section \ref{sec:web-api} (Def. \ref{def:api} -- \ref{def:complex-parameter}).
The visualisation of the Web API in Fig. \ref{fig:kMeansAPI} adopts the notation of the UML class-diagram.

\textbf{Problem description.}
Running the front-end of the app on a smartphone LG Nexus 5 (2.3 GHz CPU, 2GB RAM\footnote{\url{https://www.gsmarena.com/lg_nexus_5-5705.php}}) for small datasets ($\leq$ 1000 bidding data), Emma was happy to see that the response times were ranging from $0.15$ seconds to $34$ seconds.
However, running the app for larger datasets ($\approx$ $5000$ bidding data), she experienced the response time of $\approx$ $3.5$ minutes.
Emma could not accept this response time because Emma wants to determine her bid for a product quickly.

\textbf{Sketching the solution to problem.}
Tom observed that the response time of the app was increased due to the network latency and especially the latency of transmitting many bidding data.
To avoid this, Tom wants to augment the cloud instance of the back-end with a service instance that is deployed on the Fog (i.e., close to Emma's device).
However, it is not clear to Tom how he should implement the \textit{dynamic switching} of the front-end between the instances of the back-end.
To help Tom in implementing the dynamic switching, we propose an automated AI-based approach that learns switching decisions between the Fog and the Cloud and dynamically makes switching decisions that achieve the lowest response-time.
Before detailing our approach, we first specify below the concept of the Fog infrastructure that we adopt.


\section{Fog Infrastructure}\label{sec:infrastructure}
The Fog infrastructure usually includes three groups of devices \cite{Bonomi:2012:FCR:2342509.2342513}.
The first group encompasses mobile devices where the front-end of mobile apps has been installed.
The second group consists of edge devices used for deploying instances of the back-end of apps.
These devices provide physical machines or virtual machines for the deployment of the service back-end of the apps and are usually more powerful than those of the first group.
The devices of the first group and the second group usually communicate to each other through one-hop (or few-hops) wireless communication links.
The communication links between the front-end and the edge instances of the back-end of an app have negligible latency.
The third group comprises the virtual machines provided by powerful remote (e.g., cloud) devices.
Overall, we define the Fog infrastructure that we adopt in this work as follows.

\begin{definition}[Fog infrastructure]\label{def:fogInfrastructure}
\textit{We define the Fog infrastructure by the following tuple:
\begin{equation}I = (M_e, M_r)\end{equation}
The tuple includes the set, $M_e = \{m_{e_i}\}$, of edge machines, and the set, $M_r = \{m_{r_i}\}$, of remote machines.} \DefBox
\end{definition}

Multiple instances of the same service back-end may have been deployed in (edge and/or remote) machines of the Fog infrastructure.
These instances are accessible through different endpoint addresses.
We define the notion of the service instance in the Fog infrastructure as follows.

\begin{definition}[Service instance through the Fog]\label{def:serviceInstance}
\textit{We define an instance, $si$, of a service back-end by the following tuple:}
\begin{equation}si = (api,\ uri,\ m,\ p)\end{equation}
\textit{The tuple includes:}
\begin{itemize}
	\item \textit{the $api$ of the service (Def. \ref{def:api})}
	\item \textit{the endpoint address, $uri$, of $si$}
	\item \textit{the machine, $m$ ($m_e$ or $m_r$), where $si$ has been deployed}
	\item \textit{machine-learning model, $p$, of response time of $si$ on $m$.}
\end{itemize}
\end{definition}

\textbf{Illustrative example}.
The front-end of the auction app is installed on a resource-constrained device (esp., a smartphone).
An instance of the \texttt{\small k-means service} of the back-end of the app is deployed on a edge machine (i.e., laptop computer).
Finally, a virtual machine on the cloud is used for deploying a remote instance of the \texttt{\small k-means service}.
The details about the Fog infrastructure that we used for the app are provided in the experimental evaluation of our approach (Section \ref{sec:evaluation}).

\section{AI Self-Back-End as a Service}\label{sec:autonomousService}
Our approach extends the back-end of mobile apps so that the back-end is converted into an AI self-back-end exposed as a service.
The back-end is converted into an AI self-back-end because the back-end includes an AI-based autonomic (i.e., self-adaptive) component \cite{DBLP:journals/computer/KephartC03}.
The autonomic component dynamically predicts the response time of the edge instances and the remote instances of the service back-end and decides to invoke the instances that achieves the lowest predicted response-time.
To do so, the autonomic component could implement the classical version of the autonomic control-loop of self-adaptive software \cite{DBLP:journals/computer/KephartC03}, as described in Section \ref{sec:classical-loop}.

\subsection{Classical Version of Autonomic Control-Loop}\label{sec:classical-loop}
The classical version of the autonomic control-loop has already been implemented in our preliminary approach specified by our past work in \cite{DBLP:conf/icsoc/AthanasopoulosM19}.
This classical version of the control loop implements the Monitor-Analyze-Plan-Execute loop of self-adaptive software \cite{DBLP:journals/computer/KephartC03}.
The entry point of the control loop is the Planning mechanism that predicts the response time of service instances and selects the instance with the lowest response-time.
In particular, the Planning mechanism uses the constructed mathematical-expressions to predict the response time of service instances.
The Execution mechanism invokes the selected service-instance via using a service proxy that is included between the autonomic component and the target service-instance.
The Execution mechanism implements the service proxy.
The Monitoring mechanism records the response time of the invocations to the edge/remote service-instances.
The Analysis mechanism (re-)builds the mathematical expression of the response time of service instances.

However, as described in Section \ref{sec:SOTA}, the version of the autonomic control-loop that has been implemented in \cite{DBLP:conf/icsoc/AthanasopoulosM19} has an important restriction.
The restriction is that it captures the response time via building mathematical expressions in a manner that is \textit{tied to the independent variables} of the expressions.
Given that the construction manner of a mathematical expression is tied to its independent variables, a new mechanism should be developed for different independent variables.
To propose a generic approach that builds mathematical expressions in a manner that is not tied to specific independent variables, we propose an AI-based version of the autonomic control-loop (Section \ref{sec:AI-loop}).
In a nutshell, the AI control-loop takes any independent variable as input and fits the independent variables to monitoring data for constructing machine-learning models of the response time of service instances.

\subsection{AI Version of Autonomic Control-Loop}\label{sec:AI-loop}
The AI version of the autonomic control-loop implements a machine-learning technique (e.g., k-nearest neighbours, support vector machine, decision tree, neural network) \cite{DataMiningBook06} that is used to \textit{pre-train} machine-learning models of the response time of service instances.
In other words, the switching decision is made based on pre-trained machine-learning expressions.
To decide a service instance, the autonomic component implements an AI variation of the classical version of the autonomic control-loop that differs from the classical version in the following points (Section \ref{sec:diffs}).

\subsubsection{AI Control-Loop vs. Classical Control-Loop}\label{sec:diffs}
This variation has some differences with respect to the classical version of the control loop that is implemented in \cite{DBLP:conf/icsoc/AthanasopoulosM19}.
The first difference is that the Planning mechanism of the AI version of the control loop now works in two separate modes: an offline training mode and an online prediction mode:
\begin{enumerate}
    \item \textit{Training mode}: the mechanism offline receives training data and trains a machine-learning model of the response time of edge/remote instances.
    \item \textit{Prediction mode}: the mechanism receives an input dataset in an online way and predicts the response time of service instances via using the trained machine-learning models.
\end{enumerate}
In the classical version of the autonomic control-loop, the Planning mechanism works in a joint online training-prediction mode \cite{DBLP:conf/icsoc/AthanasopoulosM19}.

The second difference between the versions of the control loop is that the Monitoring mechanism is not needed because the response time of the available service-instances have been already been provided to the Planning mechanism in the training mode.
The third difference is that the Analysis mechanism is not also needed because the machine-learning models of response time are not reconstructed.
Finally, the fourth difference is that the Planning mechanism does not interact with the Execution mechanism when the Planning mechanism works in the training mode because no instance is invoked in the training mode.
Having detailed the differences between the two versions of the control loop, we now move on to the description of the architecture of the AI version of the autonomic control-loop (Section \ref{sec:architecture}).

\subsubsection{Architecture of AI Autonomic Control-Loop}\label{sec:architecture}
\begin{figure}
	\centering
	\includegraphics[scale=0.9]{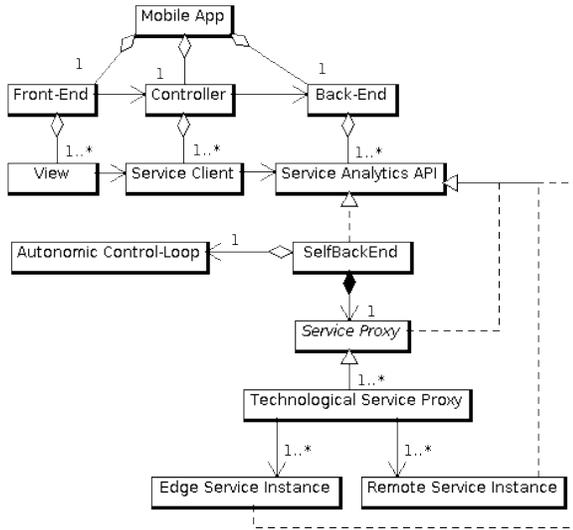}\\ 
	\caption{Mobile apps extended with AI self-back-end as a service.}
	\label{fig:proposedApplicationModel}
\end{figure}

The architecture of the AI self-back-end as a service is depicted in Fig. \ref{fig:proposedApplicationModel}.
According to this architecture, the controller of an app interacts with the AI self-back-end.
The AI self-back-end encapsulates and invokes the available edge instances and remote instances of the service back-end of the app (see the depicted instances at the bottom of Fig. \ref{fig:proposedApplicationModel}).
In other words, multiple instances of the back-end are deployed on the Fog infrastructure that is specified in Section \ref{sec:infrastructure}.

In more detail, the Web API of the AI self-back-end implements the proxy design-pattern \cite{Gamma94} because both the edge instances and the remote instances of the service back-end expose the original Web API of the back-end (see the \texttt{\small Service Proxy} abstract class in Fig. \ref{fig:proposedApplicationModel}).
The service proxy is implemented by a RESTful or a SOAP technological variations (see the \texttt{\small Technological Service Proxy} class in Fig. \ref{fig:proposedApplicationModel}).

The AI self-back-end further implements the AI version of the autonomic control-loop that trains machine-learning models of response time and self-decides at runtime the edge/remote instance that will be invoked (see the \texttt{\small Autonomic Control-Loop} class in Fig. \ref{fig:proposedApplicationModel}).
The control loop is activated when the Web API of the AI self-back-end is invoked.
The entry point of the control loop is the Planning mechanism that is invoked when the controller wishes to train machine-learning expressions or to predict the response time of service instances.
The mechanisms of the AI version of the control loop are detailed in Section \ref{sec:loop}.
Before describing the mechanisms, we illustrate the architecture of the AI self-back-end by revisiting the case study of the auction app.

\begin{figure}
	\centering
	\includegraphics[scale=0.8]{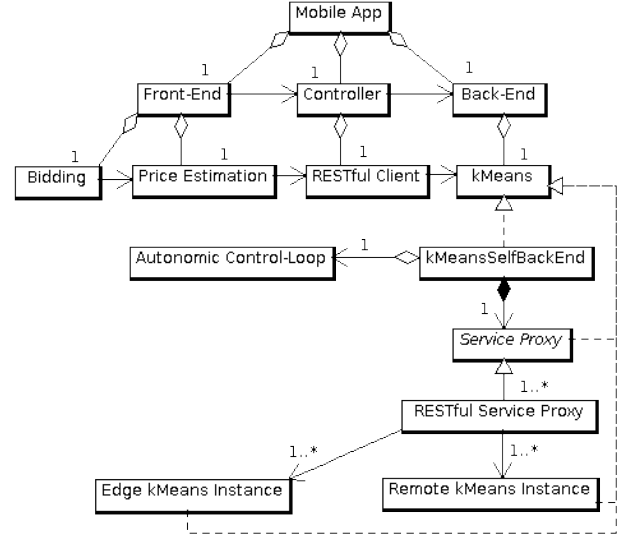}\\ 
	\caption{The auction app extended with an AI self-back-end as a service.}
	\label{fig:proposedApplicationModelAuction}
\end{figure}

\textbf{Illustrative example}.
Fig. \ref{fig:proposedApplicationModelAuction} presents the architecture of the auction app extended with an AI self-back-end as a service.
In this figure, the Web API of the service back-end corresponds to the \texttt{\small kMeans} class visualised in Fig. \ref{fig:kMeansAPI}.
The \texttt{\small kMeansSelfBackEnd} class implements the \texttt{\small kMeans} Web API and interacts with the autonomic control-loop.
The \texttt{\small kMeansSelfBackEnd} class further uses a RESTful service proxy to forward invocations to the edge/remote instances of the back-end.
In other words, all the instances of the auction back-end have been deployed as RESTful Web services (they could have been deployed as SOAP-based Web services).
Finally, the \texttt{\small kMeansSelfBackEnd} class has been exported as a RESTful service and the controller uses a RESTful client for invoking the \texttt{\small kMeansSelfBackEnd} class.

\section{AI Autonomic Control-Loop Mechanisms}\label{sec:loop}
As described in Section \ref{sec:AI-loop}, the AI autonomic control-loop includes two mechanisms out of the four mechanisms of the classical version of the autonomic control-loop: the Planning mechanism (Section \ref{sec:planning}) and the Execution mechanism (Section \ref{sec:execution}).

Both the Planning and the Execution mechanisms are deployed on an edge machine.
In this way, the network/communication latency between the controller of an app and the Planning mechanism is avoided, when the Planning mechanism is executed in prediction mode.
In a similar way, the latency between the Planning mechanism and the Execution mechanism is avoided, when the Planning mechanism interacts with the Execution mechanism.

The AI autonomic control-loop does not include further mechanisms, apart from the Planning mechanism and the Execution mechanism.
Especially, the current version of the AI autonomic control-loop \textit{does not manage} the edge/remote instances of service back-end.
In other words, the AI autonomic control-loop does not add/remove edge/remote instances or does not manage the availability/stability of the instances (leaving the dynamic management of the instances as future work).
Thus, it makes the assumption that a set of edge/remote instances is available when an app is first installed on the device of an end-user.
It further assumes that this set of edge/remote instances is not updated as the app is executed.

\subsection{Planning Mechanism}\label{sec:planning}
The Planning mechanism works in a training mode (Section \ref{sec:training-mode}) or in a prediction mode (Section \ref{sec:prediction-mode}).

\subsubsection{Training Mode}\label{sec:training-mode}
The Planning mechanism receives training/monitoring data and uses them to train one of the requested machine-learning models.
In particular, the mechanism accepts one of the following types of machine-learning models as input: k-nearest neighbours, support-vector machines, decision tree, and neural network \cite{DataMiningBook06}.
Then, the mechanism trains the requested model as follows.

\textbf{K-nearest neighbours.}
The mechanism sets the number of the neighbors to be equal to $10$ because, according to the experimental evaluation of the model of k-nearest neighbors (Section \ref{sec:evaluation}), the model achieves on average the highest accuracy for this number.

\textbf{Support-vector machines.}
According to the experimental evaluation of the model of support-vector machines (Section \ref{sec:evaluation}), the model achieves on average the highest accuracy when the mechanism sets the gamma variable to the following value:
\begin{equation} \gamma = \frac{1}{indep.\ variables + variance\ of\ indep.\ variables} \end{equation}

\textbf{Decision trees.}
The mechanism sets the three parameters (criterion, max features, min impurity split) of the model of decision-tree models as follows.
The criterion is the function to measure the quality of a split.
The mechanism calculates the value of this criterion by using the Gini impurity \cite{DataMiningBook06}.
The max number of features corresponds to the number of features to be considered in each split.
The mechanism sets the max number of features to four.
The min impurity split corresponds to the threshold for early stopping in the tree growth.
The mechanism sets the min value of the impurity split to zero.

\textbf{Neural networks.}
The mechanism implements a three-layer LeakyReLU neural network \cite{DataMiningBook06}.
The number of the neurons at the first layer equals to the number of the independent variables (see below).
For instance, the neural network for the Web API (Fig. \ref{fig:kMeansAPI}) of the auction back-end includes four neurons.
The number of the neurons at the second layer is calculated by the following expression: $\ceil{\sqrt{first\ layer\ neurons}}$.
The single neuron of the third layer corresponds to the predicted response-time.

\textbf{(In)dependent variables of the machine-learning models.}
The independent variables of all the models are related to the characteristics of the input datasets of the data-analytics algorithms.
The datasets include the values for the primitive parameters of the operations (Def. \ref{def:primitive-parameter}).
We focus on primitive parameters exclusively because primitive parameters carry on real data.
In the case of the k-means algorithm of the auction back-end, the independent variables include the following variables: the number of clusters, the number of clustering iterations, the number of data-points, and the dimension of data-points.
The dependent variable corresponds to the response time of the operations of the Web API of edge/remote instances of the back-end.

\textbf{Illustrative example}.
Revisiting Fig. \ref{fig:kMeansAPI}, the Planning mechanism accepts as input: the clusters number, $k$, the clustering iterations, $it$, the dimension, $d$, of the parameter, \texttt{\small Vector}, the length of the parameter, \texttt{\small DataPoint} (along with the values of each one of the data points).
As an example, the mechanism accepts the following dataset as input: $3$ clusters, $100$ iterations, and $1000$ data points of $3$ dimensions.

\textbf{Training machine-learning models for service instances.}
The mechanism trains a separate machine-learning model for each edge/remote instance of a back-end.

\textbf{Training a large number of service instances.}
The training mode of the mechanism is executed offline and does not affect the performance of an app.
However, the mechanism avoids a large training-time in the case of large number of available edge/remote instances.
In particular, the mechanism avoids a large training-time by training the expression of each instance in parallel.
The parallel training applies a multi-thread technique by executing one thread for the training of each edge/remote instance.
The parallel training of multiple machine-learning models is feasible because they are independent to each other.

\subsubsection{Prediction Mode}\label{sec:prediction-mode}
The Planning mechanism receives the current datasets provided from the front-end to the AI self-back-end.
Moreover, the mechanism accepts the desired kind of machine-learning model (KNN, SVM, decision tree, or neural network) as input.
To make a prediction based on a trained model, the mechanism first restores the trained model.
Then, the mechanism predicts the response times of all the service instances via using the requested machine-learning model.
The mechanism selects the instance with the lowest response-time.
Finally, the Planning mechanism interacts with the Execution mechanism to invoke the selected instance.

\textbf{Making a large number of predictions.}
The prediction mode of the Planning mechanism is executed online and affects the performance of the app.
However, the time that the mechanism needs to make the predictions for all the edge/remote instances scales in a linear way with the number of the instances.
This time is practically small because the mechanism needs a time in the order of milliseconds to make predictions for a hundred of instances.

\subsection{Execution Mechanism}\label{sec:execution}
The Execution mechanism implements a service proxy that forwards the invocations made by the front-end to the edge/remote instance selected by the Planning mechanism.
In this way, the Execution mechanism binds the front-end of the app to the back-end in a seamless way, i.e. without suspending the app execution.
The Execution mechanism provides two implementations of an abstract proxy: one implementation for RESTful Web services and one implementation for SOAP-based services.

\section{Experimental Evaluation}\label{sec:evaluation}
We evaluate the accuracy of our approach via measuring the percentages of the correct switching-decisions made by the AI self-back-end (Section \ref{sec:results}).
We further evaluate the efficiency of our approach via measuring the improvement in the response time of an app when the AI self-back-end is used (Section \ref{sec:results}).
Prior to presenting the results, we set up our experiments (Section \ref{sec:setup}).

\subsection{Experimental Setup}\label{sec:setup}
\textbf{Mobile app.}
We extended an open-source auction app\footnote{\url{https://github.com/jagmohansingh/auction-system}} with a research-prototype (implemented in Java) of the AI self-back-end that is exposed as a RESTful Web service.

\textbf{Datasets.}
We use datasets collected from publicly available UC Irvine machine-learning repository\footnote{\url{http://archive.ics.uci.edu/ml/index.php}}.
Each dataset consists of multi-dimensional data-points.
We selected datasets that can be used for online data-analytics.
In other words, we selected $11577$ datasets for which the execution time of the k-means algorithm on the used edge/remote machines is low\footnote{The datasets we use in the experiments are available here: \url{https://drive.google.com/file/d/1h9I9v_O9r03j3z-axXEmxI50RZs9F9An/view?usp=sharing}}.
In particular, we selected datasets whose data-points' number is lower than $12000$ and the data-points' dimension is at most $14$.
The overall statistics of the used datasets are presented in Table \ref{table:datasets}.
The repository provides a small number of datasets that contain a very high number of data-points.
The statistics of the datasets used in each machine-learning model are presented in Sections \ref{sec:neural-results}-\ref{sec:decision-tree-results}.

\begin{table}
\begin{center}
\caption{Overall statistics of the datasets we used in the experiments.\label{table:datasets}}
\vspace{-0.2cm}
\scalebox{1.0}{
	\begin{tabular}{r| r| r}
	    \toprule
		\textbf{Data-Points} & \textbf{Dimension of Data-Points} & \textbf{Number of Datasets} \\ \hline\hline
		$[1, 999]$ & $[3, 14]$ & $3131$ \\ 
		$[1000, 1999]$ & $[3, 14]$ & $1908$ \\ 
		$[2000, 2999]$ & $[3, 14]$ & $1916$ \\ 
	    $[3000, 3999]$ & $[3, 14]$ & $1837$ \\ 
		$[4000, 4999]$ & $[3, 12]$ & $1891$ \\ 
		$[5000, 5999]$ & $[3, 12]$ & $157$ \\ 
		$[6000, 6999]$ & $[3, 14]$ & $164$ \\ 
		$[7000, 7999]$ & $[3, 11]$ & $169$ \\ 
		$[8000, 8999]$ & $[3, 11]$ & $162$ \\ 
		$[9000, 9999]$ & $[3, 12]$ & $196$ \\ 
		$[10000, 10999]$ & $[7, 8]$ & $46$ \\ \hline
		\multicolumn{2}{r}{\textbf{Total number of the datasets:}} & $11577$ \\
	    \bottomrule
	\end{tabular}
}
\end{center}
\end{table}

\textbf{Fog infrastructure.}
The front-end of the auction app runs on a mobile phone.
We further used two laptop devices.
Each laptop provides an edge machine.
The first laptop used for deploying an edge instance of the original service back-end of the app.
The second laptop used for deploying an edge instance of the component of the AI autonomic control-loop.
The mobile phone and the laptops are connected at the same wireless local-network.
A remote instance of the original back-end is deployed on Microsoft Azure Cloud Service in the UK-south region.
All the instances of the service back-end are exposed as RESTful Web services.

\subsubsection{Accuracy Metrics}
To measure the accuracy of our approach, we first calculate the precision and the recall of each switching decision \cite{DBLP:books/aw/Baeza-YatesR99}.
In particular, we construct for each switching decision the confusion matrix demonstrated in Table \ref{table:confusion-matrix}.
\begin{table}
    \begin{center}
	\caption{The confusion matrix for calculating precision and recall.}\label{table:confusion-matrix}
	\vspace{-0.2cm}
	\scalebox{1.0}{
	\begin{tabular}{c c| c| c}
	    \toprule
        & & \multicolumn{2}{c}{\textbf{Prediction}}\\ \cline{3-4}
        & & \textit{Remote} & \textit{Edge}\\ \hline\hline
        \multicolumn{1}{c|}{\textbf{Correct}} & \textit{Remote} & True remote & False edge \\ \cline{2-4}
        \multicolumn{1}{c|}{\textbf{Decision}} & \textit{Edge} & False remote & True edge \\
		\bottomrule
	\end{tabular}}
	\end{center}
\end{table}
We then calculate the precision of all the remote (resp., edge) switching-decisions that equals to the percentage of the correct predictions over all the remote (resp., edge) switching-decisions, as defined below.
\begin{equation} remote\ precision := \frac{true\ remote}{true\ remote + false\ remote}\label{eq:remote-precision}\end{equation}
\begin{equation} edge\ precision := \frac{true\ edge}{true\ edge + false\ edge}\end{equation}
The recall of all the remote (resp., edge) switching-decisions equals to the percentage of the correct predictions over all the correct remote (resp., edge) switching-decisions and the wrong edge (resp., remote) switching-decisions, as defined below.
\begin{equation} remote\ recall := \frac{true\ remote}{true\ remote + false\ edge}\end{equation}
\begin{equation} edge\ recall := \frac{true\ edge}{true\ edge + false\ remote}\label{eq:edge-recall}\end{equation}

Finally, we measure the overall accuracy of our approach over all the switching decisions, as follows:
\begin{equation} Overall\ accuracy := \frac{correct\ switching\ decisions}{all\ switching\ decisions}\label{eq:overall-accuracy}\end{equation}

\begin{table*}
    \begin{center}
	\caption{The configuration of the AI self-back-end in the experiments to train/use various machine-learning models.}\label{table:configurations}
	\vspace{-0.2cm}
	\scalebox{1.0}{
	\begin{tabular}{r|| c| c| c| c| c| c}
	    \toprule
        & \textbf{Remote CPUs} & \textbf{Remote Memory} & \textbf{Edge CPUs} & \textbf{Edge Memory} & \textbf{Training Datasets} & \textbf{Testing Datasets}\\ \hline\hline
        \textit{Neural networks} & 4 & 8GB & 0.5 & 1GB & 578 & 925 \\
        \textit{K-nearest neighbours} & 4 & 8GB & 0.5 & 1GB & 578 & 810 \\
        \textit{Support-vector machines} & 4 & 8GB & 0.5 & 1GB & 578 & 1540 \\
        \textit{Decision trees} & 4 & 8GB & 0.5 & 1GB & 578 & 1016 \\
		\bottomrule
	\end{tabular}}
	\end{center}
\end{table*}

\subsubsection{Efficiency Metrics}
To evaluate the efficiency of our approach, we take into account the fact that an app would use only remote instances of a back-end if we do not apply our approach.
By applying our approach, there is an improvement in the response time of the app if a correct switching-decision is the invocation of an edge instance and the AI self-back-end decides the edge instance.
In this case, the response time is reduced by the difference of the response time of the app using a remote instance from the response time of the app using the edge instance.
Thus, the improvement in the response time (RTI) of the app for a single switching decision is defined below.
\begin{equation} edge\ RTI := \frac{remote\ RT - SelfBackEnd\ edge\ RT}{remote\ RT}\end{equation}
If the correct switching-decision is the invocation of a remote instance and the AI self-back-end decides the remote instance, then there is no response-time improvement because the remote instance should had by default been invoked.
We finally calculate the average improvement in the response time of an app for all the switching decisions, as follows.
\begin{equation} \overline{edge\ RTI} := \frac{\sum edge\ RTI}{edge\ switching\ decisions}\label{eq:edge-RTI}\end{equation}

\subsection{Experimental Results}\label{sec:results}
We repeat below the same set of experiments four times, one time for each one of the following machine-learning models: neural networks, k-nearest neighbours, support-vector machines, and decision trees (Sections \ref{sec:neural-results}-\ref{sec:decision-tree-results}).
To this end, we configure the AI self-back-end of the auction app to use one of the models at each time.
Table \ref{table:configurations} presents the hardware characteristics of the machines used for this experiment, along with the numbers of the training/testing datasets.
We use the same numbers of CPUs and the same amount of RAM in all the machine-learning models for uniformity reasons.
We also use the same number of training datasets for all the models.
Concerning the testing datasets, we present a different number of training datasets for each model.
The reason is because we want to present for each model, \textit{representative datasets} that show the variation of the accuracy of the model (from the smallest accuracy to the highest accuracy).
The number of the testing datasets that we present is different in each model because the accuracy of each model vary in a different manner from the other models.

\subsubsection{AI Self-Back-End Uses Neural Networks}\label{sec:neural-results}
\textbf{Experimental configuration.}
Table \ref{table:datasets1} presents the statistics of the $925$ representative testing-datasets of this experiment.
The datasets span all the available ranges of data-points and as expected, a small number of datasets contain a very high number of data-points.

\begin{table}
\begin{center}
\caption{The statistics of the testing datasets we used in neural networks.}\label{table:datasets1}
\vspace{-0.2cm}
\scalebox{1.0}{
	\begin{tabular}{r| r| r}
	    \toprule
		\textbf{Data-Points} & \textbf{Dimension of Data-Points} & \textbf{Number of Datasets} \\ \hline\hline
		$[1, 999]$ & $[3, 14]$ & $240$ \\ 
		$[1000, 1999]$ & $[3, 9]$ & $153$ \\ 
		$[2000, 2999]$ & $[3, 8]$ & $175$ \\ 
	    $[3000, 3999]$ & $[3, 11]$ & $141$ \\ 
		$[4000, 4999]$ & $[3, 11]$ & $140$ \\ 
		$[5000, 5999]$ & $[3, 12]$ & $11$ \\ 
		$[6000, 6999]$ & $[3, 8]$ & $16$ \\ 
		$[7000, 7999]$ & $[3, 8]$ & $11$ \\ 
		$[8000, 8999]$ & $[3, 8]$ & $17$ \\ 
		$[9000, 9999]$ & $[3, 8]$ & $18$ \\ 
		$[10000, 10999]$ & $8$ & $3$ \\ \hline
		\multicolumn{2}{r}{\textbf{Total number of the testing datasets:}} & $925$ \\
	    \bottomrule
	\end{tabular}
}
\end{center}
\end{table}

\begin{figure}
\begin{center}
    \scalebox{0.85}{
    \begin{tikzpicture}
    \pgfplotsset{height=4cm, width=10cm}
    \begin{groupplot}[group style={group size=1 by 1, group name=plots, horizontal sep=1.0cm, vertical sep=2.0cm}, legend to name=grouplegend, legend columns=-1, enlarge x limits=0.05]
    
    \nextgroupplot[xmin=500, xmax=1550, y label style={at={(-0.15,0.5)}}, ylabel=\textbf{Accuracy [\%]}, xlabel=\textbf{Dataset}, ymin=45, ymax=110, xtick={500,700,...,1700}, ytick={40,60,...,100}]
    \addplot+[draw=black, mark=none, line width=1pt] table [x=Dataset index, y=Accuracy, col sep=semicolon] {neural-network-testing.csv};
    
    \end{groupplot}
    \end{tikzpicture}}
\end{center}
\vspace{-0.6cm}
\caption{The overall accuracy of the AI self-back-end in neural networks.}
\label{fig:neural-network}
\end{figure}
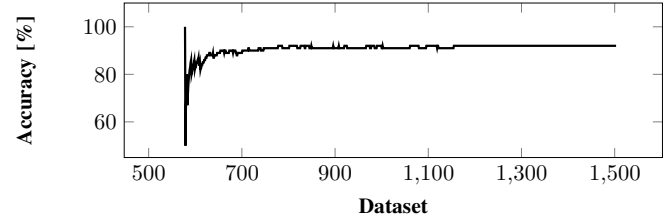

\textbf{Accuracy.}
Fig. \ref{fig:neural-network} presents the overall accuracy of the neural-network model for each testing dataset (Eq. \ref{eq:overall-accuracy}).
$925$ switching decisions made by the AI self-back-end in the prediction mode.
Please note that we present in Fig. \ref{fig:neural-network} the accuracy of the model for the datasets with IDs that are greater than $578$ because we use the first $578$ datasets for training the neural-network model.
$825$ out of $925$ switching decisions are correct decisions.
We depict at the left-hand side of the x-axis of the chart of Fig. \ref{fig:neural-network}, representative datasets for which the accuracy of the model varies/fluctuates.
We depict at the right-hand side of the x-axis, approximately $250$ representative datasets for which the accuracy of the model is stabilised.
The overall accuracy of the model is $92\%$.

\begin{table}
    \begin{center}
	\caption{The confusion matrix and the precision/recall of neural networks.}\label{table:confusion-matrix-neuron-network}
	\vspace{-0.4cm}
	\scalebox{0.97}{
	\begin{tabular}{c| c| c}
	    \toprule
        & \textbf{Remote} & \textbf{Edge}\\ \hline\hline
        \textbf{Correct Remote} & 715 & 52 \\ \hline
        \textbf{Correct Edge} & 18 & 140 \\
		\bottomrule
	\end{tabular}
	\hspace{0.1cm}
	\begin{tabular}{c| c| c}
	    \toprule
        & \textbf{Precision} & \textbf{Recall}\\ \hline\hline
        \textbf{Remote} & 98\% & 93\% \\ \hline
        \textbf{Edge} & 73\% & 89\% \\
		\bottomrule
	\end{tabular}
	}
	\end{center}
\end{table}

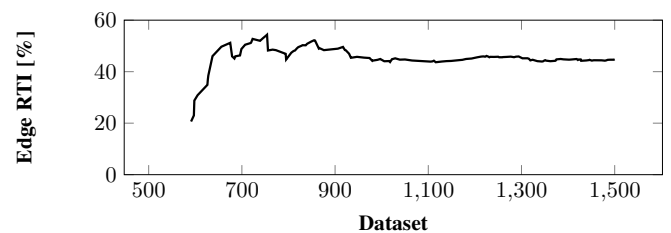
\begin{figure}
\begin{center}
    \scalebox{0.85}{
    \begin{tikzpicture}
    \pgfplotsset{height=4cm, width=10cm}
    \begin{groupplot}[group style={group size=1 by 1, group name=plots, horizontal sep=1.0cm, vertical sep=2.0cm}, legend to name=grouplegend, legend columns=-1, enlarge x limits=0.05]
    
    \nextgroupplot[xmin=500, xmax=1550, y label style={at={(-0.15,0.5)}}, ylabel=\textbf{Edge RTI [\%]}, xlabel=\textbf{Dataset}, ymin=0, ymax=60, xtick={500,700,...,1700}, ytick={0,20,...,100}]
    \addplot+[draw=black, mark=none, line width=1pt] table [x=DatasetIndex, y=Improvement, col sep=semicolon] {testing_phase_data-nn.txt};
    
    \end{groupplot}
    \end{tikzpicture}}
\end{center}
\vspace{-0.6cm}
\caption{The response-time improvement of the app in neural networks.}
\label{fig:neural-network-times}
\end{figure}

To get an insight in the switching decisions, we present in Table \ref{table:confusion-matrix-neuron-network} the confusion matrices of the switching decisions.
We further present in Table \ref{table:confusion-matrix-neuron-network} the precision/recall of the model (Eq. \ref{eq:remote-precision}-\ref{eq:edge-recall}).
We observe the AI self-back-end that is configured to use neural networks achieves high precision ($\geq73\%$) and very high recall ($\geq89\%$).

\textbf{Response-time improvement.}
We calculated the response-time improvement of the app in the edge switching-decisions for neural networks (Fig. \ref{fig:neural-network-times}).
We depict at the left-hand side of the x-axis of the chart of Fig. \ref{fig:neural-network-times}, the datasets for which the accuracy of the neural-network model varies/fluctuates.
As expected, the response-time improvement varies/fluctuates too.
We also depict at the right-hand side of the x-axis, the datasets for which the response-time improvement is stabilised (in accordance with the accuracy of the model).
The average response-time improvement in neural networks is $45\%$.

\subsubsection{AI Self-Back-End Uses K-Nearest Neighbours}
\textbf{Experimental configuration.}
Table \ref{table:datasets2} presents the statistics of the $810$ representative testing-datasets of this experiment.
The datasets span all the available ranges of data-points and as expected, a small number of datasets contain a very high number of data-points.

\begin{table}
\begin{center}
\caption{The statistics of the testing datasets we used in k-nearest neighbours.}\label{table:datasets2}
\vspace{-0.2cm}
\scalebox{1.0}{
	\begin{tabular}{r| r| r}
	    \toprule
		\textbf{Data-Points} & \textbf{Dimension of Data-Points} & \textbf{Number of Datasets} \\ \hline\hline
		$[1, 999]$ & $[3, 12]$ & $231$ \\ 
		$[1000, 1999]$ & $[3, 8]$ & $137$ \\ 
		$[2000, 2999]$ & $[3, 8]$ & $129$ \\ 
	    $[3000, 3999]$ & $[3, 14]$ & $132$ \\ 
		$[4000, 4999]$ & $[3, 12]$ & $124$ \\ 
		$[5000, 5999]$ & $[3, 8]$ & $10$ \\ 
		$[6000, 6999]$ & $[3, 7]$ & $13$ \\ 
		$[7000, 7999]$ & $[3, 11]$ & $10$ \\ 
		$[8000, 8999]$ & $[3, 11]$ & $13$ \\ 
		$[9000, 9999]$ & $[3, 8]$ & $9$ \\ 
		$[10000, 10999]$ & $8$ & $2$ \\ \hline
		\multicolumn{2}{r}{\textbf{Total number of testing datasets:}} & $810$ \\
	    \bottomrule
	\end{tabular}
}
\end{center}
\end{table}

\begin{figure}
\begin{center}
    \scalebox{0.85}{
    \begin{tikzpicture}
    \pgfplotsset{height=4cm, width=10cm}
    \begin{groupplot}[group style={group size=1 by 1, group name=plots, horizontal sep=1.0cm, vertical sep=2.0cm}, legend to name=grouplegend, legend columns=-1, enlarge x limits=0.05]
    
    \nextgroupplot[xmin=500, xmax=1400, y label style={at={(-0.15,0.5)}}, ylabel=\textbf{Accuracy [\%]}, xlabel=\textbf{Dataset}, ymin=45, ymax=110, xtick={500,800,...,1400}, ytick={40,60,...,100}]
    \addplot+[draw=black, mark=none, line width=1pt] table [x=Dataset index, y=Accuracy, col sep=semicolon] {k-nearest-neighbours-testing.csv};
    
    \end{groupplot}
    \end{tikzpicture}}
\end{center}
\vspace{-0.6cm}
\caption{Overall accuracy of the AI self-back-end in k-nearest neighbours.}
\label{fig:k-nearest-neighbours}
\end{figure}
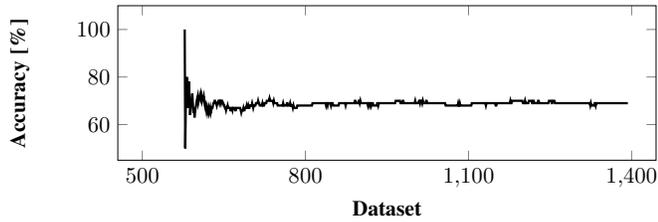

\textbf{Accuracy.}
Fig. \ref{fig:k-nearest-neighbours} presents the overall accuracy of the k-nearest-neighbours model for each testing dataset (Eq. \ref{eq:overall-accuracy}).
$810$ switching decisions made by the AI self-back-end in the prediction mode.
We present in Fig. \ref{fig:k-nearest-neighbours} the accuracy of the model for the datasets with IDs that are greater than $578$ because we use the first $578$ datasets for training the model.
$557$ out of $810$ switching decisions are correct decisions.
We depict at the left-hand side of the x-axis of the chart of Fig. \ref{fig:k-nearest-neighbours}, representative datasets for which the accuracy of the model varies/fluctuates.
We depict at the right-hand side of the x-axis, approximately $100$ representative datasets for which the accuracy of the model is stabilised.
The overall accuracy of the model is $69\%$.

To get an insight in the switching decisions, we present in Table \ref{table:confusion-matrix-k-nearest-neighbours} the confusion matrices of the switching decisions and the precision/recall of the model (Eq. \ref{eq:remote-precision}-\ref{eq:edge-recall}).
We observe the AI self-back-end that is configured to use k-nearest neighbours achieves high precision ($\geq63\%$) and medium recall ($\geq46\%$).

\begin{table}
    \begin{center}
	\caption{The confusion matrix and the precision/recall of k-nearest neighbours.}\label{table:confusion-matrix-k-nearest-neighbours}
	\vspace{-0.4cm}
	\scalebox{0.97}{
	\begin{tabular}{c| c| c}
	    \toprule
        & \textbf{Remote} & \textbf{Edge}\\ \hline\hline
        \textbf{Correct Remote} & 813 & 85 \\ \hline
        \textbf{Correct Edge} & 168 & 144 \\
		\bottomrule
	\end{tabular}
	\hspace{0.1cm}
	\begin{tabular}{c| c| c}
	    \toprule
        & \textbf{Precision} & \textbf{Recall}\\ \hline\hline
        \textbf{Remote} & 71\% & 83\% \\
        \textbf{Edge} & 63\% & 46\% \\
		\bottomrule
	\end{tabular}
	}
	\end{center}
\end{table}

\begin{figure}
\begin{center}
    \scalebox{0.85}{
    \begin{tikzpicture}
    \pgfplotsset{height=4cm, width=10cm}
    \begin{groupplot}[group style={group size=1 by 1, group name=plots, horizontal sep=1.0cm, vertical sep=2.0cm}, legend to name=grouplegend, legend columns=-1, enlarge x limits=0.05]
    
    \nextgroupplot[xmin=500, xmax=1400, y label style={at={(-0.15,0.5)}}, ylabel=\textbf{Edge RTI [\%]}, xlabel=\textbf{Dataset}, ymin=0, ymax=30, xtick={500,800,...,1400}, ytick={0,10,...,100}]
    \addplot+[draw=black, mark=none, line width=1pt] table [x=DatasetIndex, y=Improvement, col sep=semicolon] {testing_phase_data-knn.txt};
    
    \end{groupplot}
    \end{tikzpicture}}
\end{center}
\vspace{-0.6cm}
\caption{The response-time improvement in k-nearest neighbours.}
\label{fig:k-nearest-neighbours-times}
\end{figure}
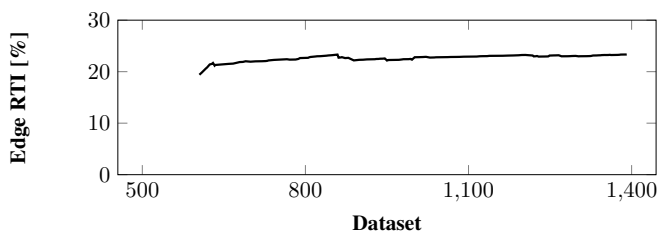

\textbf{Response-time improvement.}
We calculated the response-time improvement of the app in the edge switching-decisions for k-nearest neighbours (Fig. \ref{fig:k-nearest-neighbours-times}).
We depict at the left-hand side of the x-axis of Fig. \ref{fig:k-nearest-neighbours-times}, the datasets for which the accuracy of the model of the k-nearest neighbours varies/fluctuates.
We also depict at the right-hand side of the x-axis, the datasets for which the accuracy of the model is stabilised.
As expected, the response-time improvement increases as the accuracy is stabilised.
The average response-time improvement in k-nearest neighbours is $23\%$.

\subsubsection{AI Self-Back-End Uses Support-Vector Machines}
\textbf{Experimental configuration.}
Table \ref{table:datasets3} presents the $1540$ representative testing-datasets of this experiment.
The datasets span all the available ranges of data-points and as expected, a small number of datasets contain a very high number of data-points.

\begin{table}
\begin{center}
\caption{The statistics of the testing datasets in support-vector machines.}\label{table:datasets3}
\vspace{-0.2cm}
\scalebox{1.0}{
	\begin{tabular}{r| r| r}
	    \toprule
		\textbf{Data-Points} & \textbf{Dimension of Data-Points} & \textbf{Number of Datasets} \\ \hline\hline
		$[1, 999]$ & $[3, 14]$ & $435$ \\ 
		$[1000, 1999]$ & $[3, 12]$ & $255$ \\ 
		$[2000, 2999]$ & $[3, 9]$ & $244$ \\ 
	    $[3000, 3999]$ & $[3, 14]$ & $251$ \\ 
		$[4000, 4999]$ & $[3, 12]$ & $232$ \\ 
		$[5000, 5999]$ & $[3, 8]$ & $26$ \\ 
		$[6000, 6999]$ & $[3, 8]$ & $17$ \\ 
		$[7000, 7999]$ & $[3, 11]$ & $27$ \\ 
		$[8000, 8999]$ & $[3, 8]$ & $20$ \\ 
		$[9000, 9999]$ & $[3, 8]$ & $28$ \\ 
		$[10000, 10999]$ & $8$ & $5$ \\ \hline
		\multicolumn{2}{r}{\textbf{Total number of the testing datasets:}} & $1540$ \\
	    \bottomrule
	\end{tabular}
}
\end{center}
\end{table}

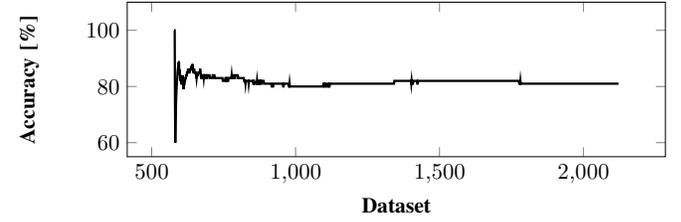
\begin{figure}
\begin{center}
    \scalebox{0.85}{
    \begin{tikzpicture}
    \pgfplotsset{height=4cm, width=10cm}
    \begin{groupplot}[group style={group size=1 by 1, group name=plots, horizontal sep=1.0cm, vertical sep=2.0cm}, legend to name=grouplegend, legend columns=-1, enlarge x limits=0.05]
    
    \nextgroupplot[xmin=500, xmax=2200, y label style={at={(-0.15,0.5)}}, ylabel=\textbf{Accuracy [\%]}, xlabel=\textbf{Dataset}, ymin=55, ymax=110, xtick={500,1000,...,2500}, ytick={40,60,...,100}]
    \addplot+[draw=black, mark=none, line width=1pt] table [x=Dataset index, y=Accuracy, col sep=semicolon] {support-vector-machine-testing.csv};
    
    \end{groupplot}
    \end{tikzpicture}}
\end{center}
\vspace{-0.6cm}
\caption{The overall accuracy in support-vector machines.}
\label{fig:support-vector-machine}
\end{figure}

\textbf{Accuracy.}
Fig. \ref{fig:support-vector-machine} presents the overall accuracy of the model of the support-vector machines for each testing dataset (Eq. \ref{eq:overall-accuracy}).
$1540$ switching decisions made by the AI self-back-end in the prediction mode.
We present in Fig. \ref{fig:support-vector-machine} the accuracy of the model for the datasets with IDs that are greater than $578$ because we use the first $578$ datasets for training the model.
$1244$ out of $1540$ switching decisions are correct decisions.
We depict at the left-hand side of the x-axis of the chart of Fig. \ref{fig:support-vector-machine}, representative datasets for which the accuracy of the model varies/fluctuates.
We depict at the right-hand side of the x-axis, approximately $30$ representative datasets for which the accuracy of the model is stabilised.
The overall accuracy of the model is $81\%$.

To get an insight in the switching decisions, we present in Table \ref{table:confusion-matrix-support-vector-machine} the confusion matrices of the switching decisions for this experiment.
We further present in Table \ref{table:confusion-matrix-support-vector-machine} the precision and the recall of the model (Eq. \ref{eq:remote-precision}-\ref{eq:edge-recall}).
We observe in the results that the AI self-back-end that is configured to use support-vector machines achieves high precision ($\geq81\%$) and low recall ($\geq9\%$).

\begin{table}[h]
    \begin{center}
	\caption{Confusion matrix and precision/recall of support-vector machines.}\label{table:confusion-matrix-support-vector-machine}
	\vspace{-0.4cm}
	\scalebox{0.97}{
	\begin{tabular}{c| c| c}
	    \toprule
        & \textbf{Remote} & \textbf{Edge}\\ \hline\hline
        \textbf{Correct Remote} & 1215 & 3 \\ \hline
        \textbf{Correct Edge} & 293 & 29 \\
		\bottomrule
	\end{tabular}
	\hspace{0.1cm}
	\begin{tabular}{c| c| c}
	    \toprule
        & \textbf{Precision} & \textbf{Recall}\\ \hline\hline
        \textbf{Remote} & 81\% & 100\% \\
        \textbf{Edge} & 91\% & 9\% \\
		\bottomrule
	\end{tabular}
	}
	\end{center}
\end{table}

\begin{figure}
\begin{center}
    \scalebox{0.85}{
    \begin{tikzpicture}
    \pgfplotsset{height=4cm, width=10cm}
    \begin{groupplot}[group style={group size=1 by 1, group name=plots, horizontal sep=1.0cm, vertical sep=2.0cm}, legend to name=grouplegend, legend columns=-1, enlarge x limits=0.05]
    
    \nextgroupplot[xmin=500, xmax=2200, y label style={at={(-0.15,0.5)}}, ylabel=\textbf{Edge RTI [\%]}, xlabel=\textbf{Dataset}, ymin=0, ymax=100, xtick={500,1000,...,2500}, ytick={0,20,...,100}]
    \addplot+[draw=black, mark=none, line width=1pt] table [x=DatasetIndex, y=Improvement, col sep=semicolon] {testing_phase_data-svm.txt};
    
    \end{groupplot}
    \end{tikzpicture}}
\end{center}
\vspace{-0.6cm}
\caption{The response-time improvement in support-vector machines.}
\label{fig:support-vector-machine-times}
\end{figure}
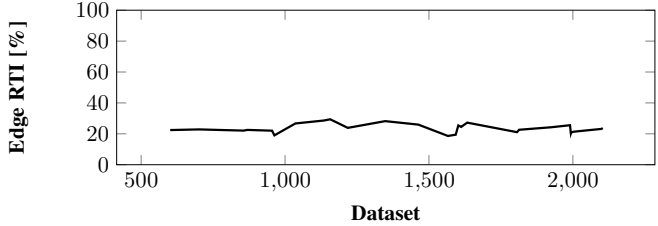

\textbf{Response-time improvement.}
We calculated the response-time improvement of the app in the edge switching-decisions for support-vector machines (Fig. \ref{fig:support-vector-machine-times}).
We depict at the left-hand side of the x-axis of Fig. \ref{fig:support-vector-machine-times}, the datasets for which the accuracy of the model of support-vector machines varies/fluctuates.
As expected, the response-time improvement varies/fluctuates for the first datasets, but it is stabilised for the datasets for which the accuracy is stabilised too.
The average response-time improvement in support-vector machines is $20\%$.

\subsubsection{AI Self-Back-End Uses Decision Trees}\label{sec:decision-tree-results}
\textbf{Experimental configuration.}
Table \ref{table:datasets4} presents the $1016$ representative testing-datasets of this experiment.
The datasets span all the available ranges of data-points and as expected, a small number of datasets contain a very high number of data-points.

\begin{table}
\begin{center}
\caption{The statistics for the testing datasets we used in decision trees.}\label{table:datasets4}
\vspace{-0.2cm}
\scalebox{1.0}{
	\begin{tabular}{r| r| r}
	    \toprule
		\textbf{Data-Points} & \textbf{Dimension of Data-Points} & \textbf{Number of Datasets} \\ \hline\hline
		$[1, 999]$ & $[3, 14]$ & $284$ \\ 
		$[1000, 1999]$ & $[3, 12]$ & $169$ \\ 
		$[2000, 2999]$ & $[3, 9]$ & $164$ \\ 
	    $[3000, 3999]$ & $[3, 11]$ & $163$ \\ 
		$[4000, 4999]$ & $[3, 12]$ & $149$ \\ 
		$[5000, 5999]$ & $[3, 4]$ & $14$ \\ 
		$[6000, 6999]$ & $[3, 8]$ & $14$ \\ 
		$[7000, 7999]$ & $[3, 11]$ & $20$ \\ 
		$[8000, 8999]$ & $[3, 8]$ & $14$ \\ 
		$[9000, 9999]$ & $[3, 8]$ & $19$ \\ 
		$[10000, 10999]$ & $8$ & $6$ \\ \hline
		\multicolumn{2}{r}{\textbf{Total number of testing datasets:}} & $1016$ \\
	    \bottomrule
	\end{tabular}
}
\end{center}
\end{table}

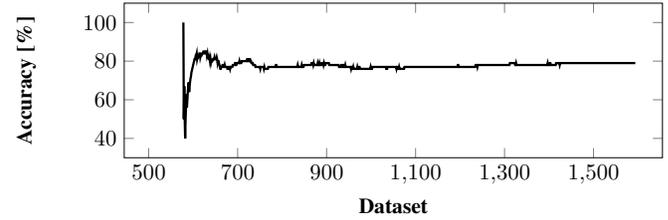
\begin{figure}
\begin{center}
    \scalebox{0.85}{
    \begin{tikzpicture}
    \pgfplotsset{height=4cm, width=10cm}
    \begin{groupplot}[group style={group size=1 by 1, group name=plots, horizontal sep=1.0cm, vertical sep=2.0cm}, legend to name=grouplegend, legend columns=-1, enlarge x limits=0.05]
    
    \nextgroupplot[xmin=500, xmax=1600, y label style={at={(-0.15,0.5)}}, ylabel=\textbf{Accuracy [\%]}, xlabel=\textbf{Dataset}, ymin=30, ymax=110, xtick={500,700,...,1600}, ytick={40,60,...,100}]
    \addplot+[draw=black, mark=none, line width=1pt] table [x=Dataset index, y=Accuracy, col sep=semicolon] {decision-tree-testing.csv};
    
    \end{groupplot}
    \end{tikzpicture}}
\end{center}
\vspace{-0.6cm}
\caption{The overall accuracy of the AI self-back-end in decision trees.}
\label{fig:decision-tree}
\end{figure}

\textbf{Accuracy.}
Fig. \ref{fig:decision-tree} presents the overall accuracy of the decision-tree model for each testing dataset (Eq. \ref{eq:overall-accuracy}).
$1016$ switching decisions made by the AI self-back-end in the prediction mode.
We present in Fig. \ref{fig:decision-tree} the accuracy of the model for the datasets with IDs that are greater than $578$ because we use the first $578$ datasets for training the model.
$804$ out of $1016$ switching decisions are correct decisions.
We depict at the left-hand side of the x-axis of the chart of Fig. \ref{fig:decision-tree}, representative datasets for which the accuracy of the model varies/fluctuates.
We depict at the right-hand side of the x-axis of the same chart, approximately $200$ representative datasets for which the accuracy of the model is stabilised.
The overall accuracy of the model is $81\%$.

To get an insight in the switching decisions, we present in Table \ref{table:confusion-matrix-decision-tree} the confusion matrices of the switching decisions and the precision/recall of the model (Eq. \ref{eq:remote-precision}-\ref{eq:edge-recall}).
We observe the AI self-back-end that is configured to use decision trees achieves medium precision ($\geq54\%$) and recall ($\geq68\%$).

\begin{table}[h]
    \begin{center}
	\caption{The confusion matrix and the precision/recall of decision trees.}\label{table:confusion-matrix-decision-tree}
	\vspace{-0.4cm}
	\scalebox{0.97}{
	\begin{tabular}{c| c| c}
	    \toprule
        & \textbf{Remote} & \textbf{Edge}\\ \hline\hline
        \textbf{Correct Remote} & 645 & 136 \\ \hline
        \textbf{Correct Edge} & 76 & 159 \\
		\bottomrule
	\end{tabular}
	\hspace{0.1cm}
	\begin{tabular}{c| c| c}
	    \toprule
        & \textbf{Precision} & \textbf{Recall}\\ \hline\hline
        \textbf{Remote} & 89\% & 83\% \\
        \textbf{Edge} & 54\% & 68\% \\
		\bottomrule
	\end{tabular}
	}
	\end{center}
\end{table}

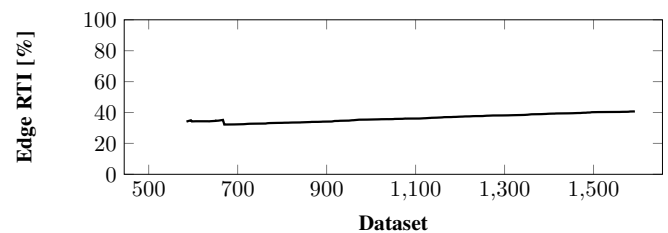
\begin{figure}
\begin{center}
    \scalebox{0.85}{
    \begin{tikzpicture}
    \pgfplotsset{height=4cm, width=10cm}
    \begin{groupplot}[group style={group size=1 by 1, group name=plots, horizontal sep=1.0cm, vertical sep=2.0cm}, legend to name=grouplegend, legend columns=-1, enlarge x limits=0.05]
    
    \nextgroupplot[xmin=500, xmax=1600, y label style={at={(-0.15,0.5)}}, ylabel=\textbf{Edge RTI [\%]}, xlabel=\textbf{Dataset}, ymin=0, ymax=100, xtick={500,700,...,1600}, ytick={0,20,...,100}]
    \addplot+[draw=black, mark=none, line width=1pt] table [x=DatasetIndex, y=Improvement, col sep=semicolon] {testing_phase_data-dt.txt};
    
    \end{groupplot}
    \end{tikzpicture}}
\end{center}
\vspace{-0.6cm}
\caption{The response-time improvement in decision trees.}
\label{fig:decision-tree-times}
\end{figure}

\textbf{Response-time improvement.}
We calculated the response-time improvement of the app in the edge switching-decisions for decision trees (Fig. \ref{fig:decision-tree-times}).
We depict at the left-hand side of the x-axis of Fig. \ref{fig:decision-tree-times}, the datasets for which the accuracy of the model of decision trees varies/fluctuates.
We also depict at the right-hand side of the x-axis, the datasets for which the accuracy of the model is stabilised.
As expected, the response-time improvement increases as the accuracy is stabilised.
The average response-time improvement in decision trees is $37\%$.

\subsubsection{Overall Comparison of the Experimental Results}
We compare in Table \ref{table:comparison} the machine-learning models we used with respect to their overall accuracy and the response-time improvement.
We observe the k-nearest neighbours model achieves the lowest accuracy, medium accuracy, and medium precision/recall.
The neural-network model achieves the highest overall accuracy, the highest precision/recall, and the highest response-time improvement.
The second better model is the decision-tree model that achieves high overall accuracy, the second highest response-time improvement, but medium precision/recall.
The support-vector-machine model has the same overall accuracy with the decision-tree model, higher precision than the precision of the decision-tree model, but it achieves very low recall and the lowest response-time improvement.

\begin{table}
    \begin{center}
	\caption{Overall comparison of the experimental results.}\label{table:comparison}
	\vspace{-0.2cm}
	\scalebox{1.0}{
	\begin{tabular}{r| c| c| c| c}
	    \toprule
        & \textbf{Accuracy} & \textbf{Precision} & \textbf{Recall} & \textbf{RTI} \\ \hline\hline
        \textit{Neural networks} & 92\% & 70\% & 89\% & 45\%\\
        \textit{K-nearest neighbours} & 69\% & 63\% & 46\% & 23\%\\
        \textit{Support-vector machines} & 81\% & 81\% & 9\% & 20\%\\
        \textit{Decision trees} & 81\% & 54\% & 68\% & 37\%\\
		\bottomrule
	\end{tabular}}
	\end{center}
\end{table}

\section{Conclusions \& Future Work}\label{sec:ConclusionsAndFutureWork}
We contributed with the specification of the architecture and the mechanisms of the AI self-back-end for edge/remote instances.
We evaluated the accuracy and efficiency of the AI self-back-end in four machine-learning models.
The evaluation results showed that the neural-network model has the highest overall accuracy and the highest precision and recall.

A future research-direction is to extend our approach to use further categories of monitoring data (e.g., memory consumption).
Another direction is to consider the synchronisation time (spent by the edge/remote instances for storing datasets) in the training of machine-learning models.
A final direction is to enhance the AI self-back-end with the capability to dynamically (de-)register service instances that are (not) available on the Fog and the Cloud.

\ifCLASSOPTIONcaptionsoff
    \newpage
\fi

\bibliographystyle{IEEEtran}
\bibliography{Arxiv.org-ACCEPTED/main-arxiv}

\begin{IEEEbiographynophoto}{Dionysis Athanasopoulos}
is a lecturer with the school of Electronics, Electrical Engineering \& Computer Science of Queen's University Belfast, UK. His research interests mainly include service-oriented architecture, Fog computing, self-adaptive software, data engineering, and software design principles \& patterns. Contact him at \texttt{D.Athanasopoulos@qub.ac.uk}.
\end{IEEEbiographynophoto}

\begin{IEEEbiographynophoto}{Dewei Liu}
received his BEng with first-class honours in Computer Science (2020) from Queen's University Belfast, UK. He is now pursuing an MSc in machine learning and data science at Imperial College London, UK. His main field of interest is computational statistics and data analysis. Contact him at \texttt{dliu08@qub.ac.uk}.
\end{IEEEbiographynophoto}

\end{document}